# Real Time Video Analysis using Smart Phone Camera for Stroboscopic Image

Somnath Mukherjee, Kritikal Solutions Pvt. Ltd. (India); Soumyajit Ganguly, International Institute of Information Technology (India)

Email:- somnath.7.mukherjee@gmail.com, soumyajit.ganguly@ymail.com

## Abstract

Motion capturing and there by segmentation of the motion of any moving object from a sequence of continuous images or a video is not an exceptional task in computer vision area. Smart-phone camera application is an added integration for the development of such tasks and it also provides for a smooth testing. A new approach has been proposed for segmenting out the foreground moving object from the background and then masking the sequential motion with the static background which is commonly known as stroboscopic image. In this paper the whole process of the stroboscopic image construction technique has been clearly described along with some necessary constraints which is due to the traditional problem of estimating and modeling dynamic background changes. The background subtraction technique has been properly estimated here and number of sequential motion have also been calculated with the correlation between the motion of the object and its time of occurrence. This can be a very effective application that can replace the traditional stroboscopic system using high end SLR cameras, tripod stand, shutter speed control and position etc.
.

## 1. Introduction

A stroboscope is an instrument which is used to project any moving object to be slowly moving or stationary or it may be define as an instrument for observing moving bodies by making them visible intermittently and thereby giving them the optical illusion of being stationary. A stroboscope may operate by illuminating the object with perfect flashes of light or by imposing an intermittent shutter between the viewer and the object. Stroboscopes are used to measure the speed of rotation or frequency of vibration of a mechanical part or system. This is belongs to the photographic world which is described in [1] as a photographic feature for construction strobes of the motion of a moving object. Traditional stroboscopic techniques in the photographic as well as filmy world generally have concentrated on simply opening the shutter at the beginning of the action and closing it at the end and recording the moving subject during the process. It becomes quickly evident that this approach has limits in terms of the length of time during which any given can be recorded because if the time is extended too far, too many images will superimpose on each other and it becomes impossible to determine the development or sequence of the action being investigated or visualized. To overcome this problem this approach can be applied instead of camera controlling where the video of the moving object or the cyclically moving in a static background is required to acquire the strobes of the sequential motion. Addition or removing object from a static background images can be estimated only by the help of a statistical model of the background scene where any intruding any object can be detected and estimated by changing of image pixel which are completely different with the statistical model of the scene, this particular functionality is known as background subtraction. The background model scene has a fixed probability density function(pdf) over each image pixel. This pixels have a correlation in between each pdf and it's description to the background model. This background scene describes the static nature when there is no abrupt changes of he image that can be only happened due to intruding new object into the scene. The pixel intensity values and it's corresponding variance shows a significant change during this time and it can be described as simple as Gaussian model [2] and also described as complex nature using Gaussian Mixture Model (GMM)[3].In reference paper [4] there are improved approach of GMM for background segmentation and also it has a wide range of optimization for development purpose Real time moving object segmentation in image sequence can be found more robustly in [5]. Adaptive Gaussian mixture model have been used along with a common optimization scheme to fit the Gaussian mixture model ,which is Expectation Maximization (EM) algorithm here. The EM algorithm is an iterative method that guarantees to converge to a local maximum in a search space. Due to the space-time requirements in modeling each pixel for the background image, an online EM algorithm is required. Many online EM algorithms have been introduced for this development also. A traditional stroboscopic image by High End DSLR traditional camera can be found in figure -2(a).

## 1.1. Related Research Work

In this section we review the most important motion segmentation categories. Frame subtraction from a continuous video is one of the simplest and most used technique for detecting changes of the background. Scene. It consists in thresholding whether it is local or global over the intensity difference of two frames pixel by pixel. The result is a coarse map of the temporal changes Despite its simplicity, this technique cannot be used in its basic version because it is really sensitive to noise. Moreover, when the camera is moving the whole image is changing and, if the frame rate is not high enough, the result would not provide any useful information. However, there are a few techniques based on this idea. The key point is to compute a rough map of the changing areas and for each blob to extract spatial or temporal information in order to track the region. Usually different strategies to make the algorithm more robust against noise and light changes are also used. Examples of this technique can be found in [10,11,12,13]. Statistic theory is widely used in the motion segmentation field. In fact, motion segmentation can be seen as a classification problem where each pixel has to be classified as background or foreground. In reference paper [15] there are mentioned various types of motion detection from where it can very easy to develop stroboscopic image construction in various angle of deviation.MAP solution has been included using Minimum Message Length criterion in [15]. Background subtraction using a static camera on a scene is also described clearly in [7] where Adaptive and Gaussian mixture model have been used for proper running average calculation between the consecutive sequence of image . A detailed review of motion segmentation can be found in [9] and also in [14] there are so many techniques have been incorporated also.

## 1.2. Background Subtraction using improved Gaussian Mixture Model(GMM)

Background subtraction using image pixel intensity variation is basically involves a certain decision about a pixel whether it belongs to a background (BG) or in foreground object (FG). In [1] there have been discussed about the Bayesian decision R for this where $\vec{x}^{(t)}$ is denoted by the colour space in RGB mode.

$$R = \frac{p(BG|\vec{x}^{(t)})}{p(FG|\vec{x}^{(t)})} = \frac{p(\vec{x}^{(t)}|BG)p(BG)}{p(\vec{x}^t|FG)p(FG)} \quad (1)$$

According to this result we can decide the image pixel belongs to the background (BG) if :

$$p(\vec{x}^{(t)}\}BG) > C_{thrsh} \quad (2)$$

Where $C_{thrsh}$ is a threshold value depends on $P(\vec{x}^t|FG)$. In general, the illumination of the light in the scene could change gradually like in out door scene or suddenly like switching any light in a black room Any new rigid object could be brought into the scene or any kind of removing present object from it. In order to estimate to changes an update the training set by adding new samples and discarding the old ones. According to reference of [6] the improved Gaussian mixture model can be represented with M components :

$$\hat{p}(\vec{x}|X_T, BG+FG) = \sum_{m=1}^{M} \hat{\pi}_m N(\vec{x}; \vec{\mu}_m, \hat{\sigma}^2_m I) \quad (3)$$

Where … $\vec{\mu}_m$ are the estimates of the means and… $\hat{\sigma}^2_m$ are the estimates of the variances that describe the Gaussian components.

## 1.3. Motion Segmentation & Alignment For Stroboscopic Image

Assuming the environmental conditions are well suited as mentioned in [6] we start recording a video for stroboscopic image . The dynamic subject which involves the motion can be present at the initial frame being captured or not be present. It hardly affects the output. The typical record time is set to be about Time-5 seconds and video frame extraction rate will be like 25 fps**.** For the entire duration of this video we perform a background modeling of the scene [6]. This algorithm takes into account subtle changes in the background and eliminates them. Examples of the subtle changes might be minor motion caused due to the shaking of hand during video shoot, branches and leaves of a tree rustling in the wind during outdoor use. This background modeling process is run-time during the video shoot. The processing power of smart phones nowadays can easily handle the processing in live video. We are also reducing the resolution of the frames before feeding them to the modeling process. The output of this stage is an image which is the background model.Next we perform background subtraction, we already have the background image and we subtract it from the input video on a frame by frame basis. After this subtraction phase, we threshold the result to get a binary image as output per frame (Figure -2(b)). For the thresholding, we use a histogram based method [3]. After we have the binary images, we perform some morphological noise removal techniques [9]. The effective result should be a white connected blob in a black image (Figure -2(b)). This white blob signifies the region in which motion was present in the particular frame. We now have an estimate of the subject in a per frame basis. Computing the central moments of these binary images would tell us the respective spatial positions of the subject as it moved along in the video. We can now combine spatial and temporal techniques to combine these blobs into a single image. Typically a 5 second video will have about 125 frames. We can tune the above techniques to get about 5 - 10 separate spatial locations of the subject and we also know the exact frame or time sequence of these spatial locations. Which of the frames to chose from the input video is determined by the spatial location of the blob. Thus we just copy over the pixels from the input video to these white blobs in the respective regions. The rest black portions are just replaced by our pre-computed background model. The final result of stroboscopic image is described in figure -2(c&d)

### 1.3. Proposed Algorithm in Flow

The whole process of the algorithmic way has been illustrated in below figure(figure-1) that can help to understand the step by step process for making a stroboscopic image.

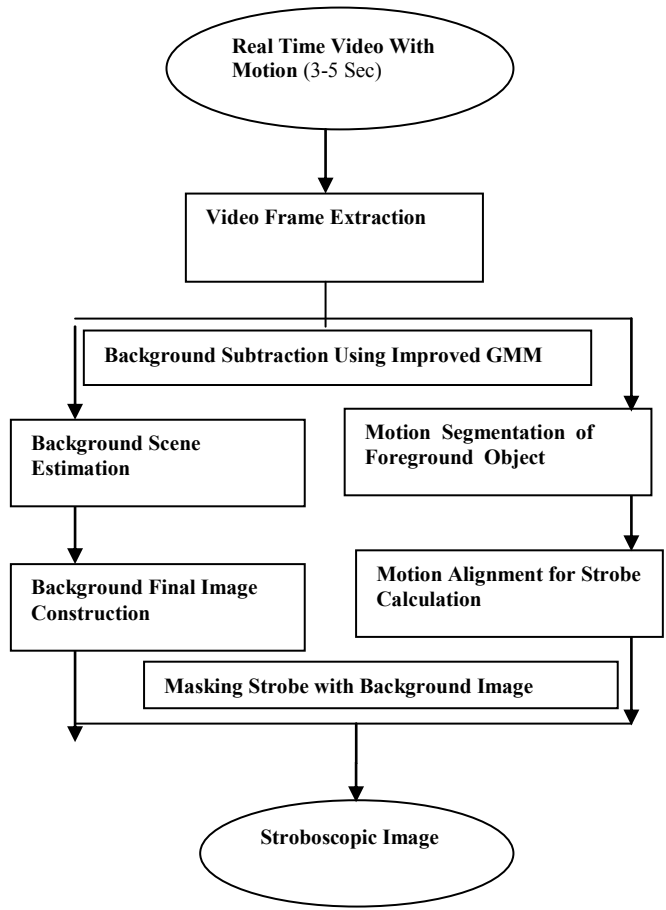

Figure 1: Demonstration of the flow chart of the algorithm step by step .

### 1.4. Discussion for some of technical difficulties and future research to make this system more robust

Ideally the user satisfaction should meet with technical feasibility of any kind of application development basically where the end user will the smart phone user. A new approach has been proposed for segmenting the foreground moving object from the background and then masking the sequential motion with the static background which is commonly known as stroboscopic image. In this approach the whole process of the stroboscopic image construction technique (current approach ) has been clearly described along with some necessary constraints which is due to the traditional problem of estimating and modeling dynamic background changes. The background subtraction technique has been properly estimated here and number of sequential motion have also been calculated taking the information from the correlation between the motion of the object and its time of occurrence. There are some provision for further research work to make this system to an advance and robust mentioned in the previous section and which is still now is not available in the current research work of the computer vision area. There are also some other provisions to make this system more advance like to incorporate the relative motion capturing robustly.

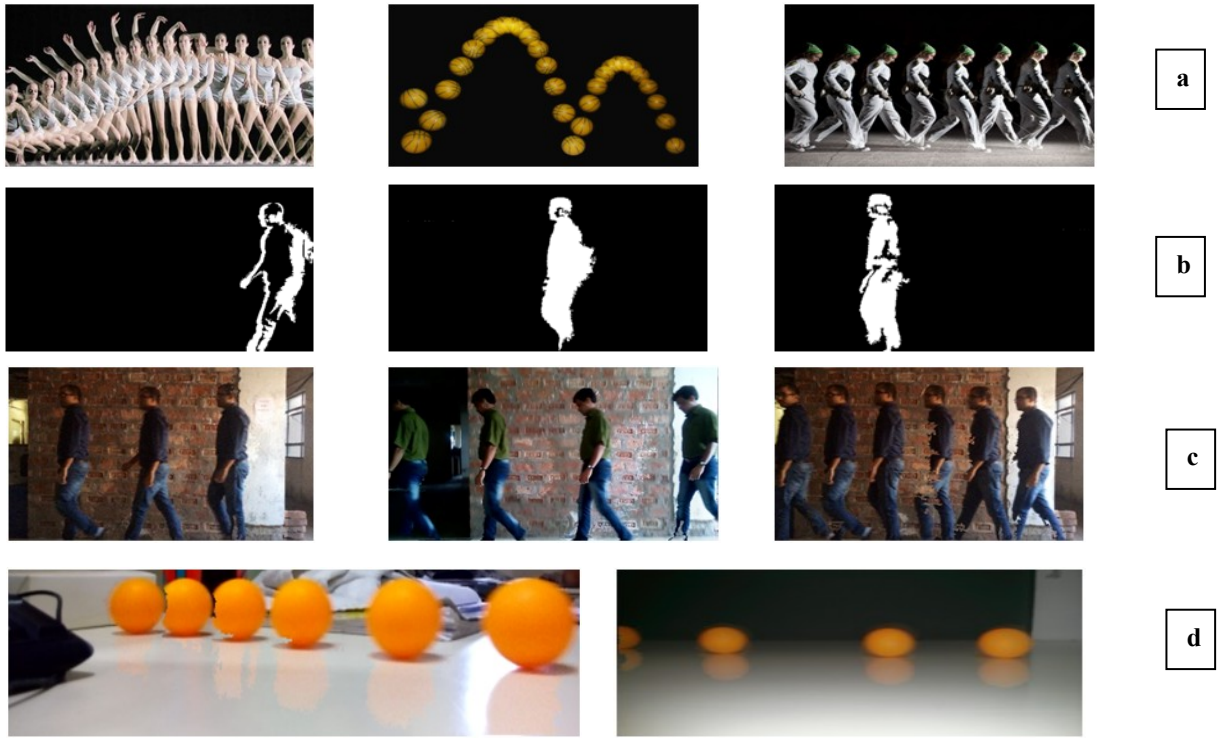

Figure 2: Stroboscopic Image (a-traditional stroboscopic image, b- Motion captured and segmentation by the proposed algorithm, c & d - experiential stroboscopic image construction after final result)

## 2. Conclusion

There are various scope for improvement of our proposed approach for motion segmentation like in [8] motion segmentation procedure have been discussed using changing with optical flow detection where clustering and grouping of motion vector so that a motion field or a particular position of motion occurrence is segmented into regions of different motion. Another way  also like estimation of global and local motion and there by differentiation of them can be a very good and adaptive approach for proper motion analysis and capturing the sequence of object strobe [8]. An application in Android platform has been released using our proposed algorithm in stroboscopic image construction. It has been released under  free to download and installation from Google Ply Store whether any one can test and provide feed back.